\pdfoutput=1
\documentclass[journal]{IEEEtran}

\usepackage{amsmath,amsfonts}
\usepackage{algorithmic}
\usepackage{algorithm}
\usepackage{array}
\usepackage[caption=false,font=normalsize,labelfont=sf,textfont=sf]{subfig}
\usepackage{textcomp}
\usepackage{stfloats}
\usepackage{url}
\usepackage{verbatim}
\usepackage{graphicx}
\usepackage{cite}
\usepackage{multirow}
\usepackage{listings}
\usepackage{xcolor}
\usepackage{comment}

\usepackage{placeins} 

\begin{document}

\title{PerNodeDrop: A Method Balancing Specialized Subnets and Regularization in Deep Neural Networks}

\author{
    Gelesh~G.~Omathil, %
    
    \thanks{
        Gelesh~G.~Omathil is a Research Scholar with the Department of Computer Science, 
        Christ University, Bangalore, India (e-mail: gelesh.omathil@res.christuniversity.in).
    }%

    \thanks{
        A provisional patent related to the PerNodeDrop regularization method 
        has been filed by Chemophilic Data Sage.
    }
}

\maketitle

\maketitle

\begin{abstract}
Deep neural networks possess strong representational capacity yet remain vulnerable to overfitting, primarily because neurons tend to co-adapt in ways that, while capturing complex and fine-grained feature interactions, also reinforce spurious and non-generalizable patterns that inflate training performance but reduce reliability on unseen data. Noise-based regularizers such as Dropout and DropConnect address this issue by injecting stochastic perturbations during training, but the noise they apply is typically uniform across a layer or across a batch of samples, which can suppress both harmful and beneficial co-adaptation.

This work introduces PerNodeDrop, a lightweight stochastic regularization method. It applies per-sample, per-node perturbations to break the uniformity of the noise injected by existing techniques, thereby allowing each node to experience input-specific variability. Hence, PerNodeDrop preserves useful co-adaptation while applying regularization. This narrows the gap between training and validation performance and improves reliability on unseen data, as evident from the experiments.

Although superficially similar to DropConnect, PerNodeDrop operates at the sample level. It drops weights at the sample level, not the batch level. An expected-loss analysis formalizes how its perturbations attenuate excessive co-adaptation while retaining predictive interactions. Empirical evaluations on vision, text, and audio benchmarks indicate improved generalization relative to the standard noise-based regularizer.

\end{abstract}

\section{Introduction}
\IEEEPARstart{D}{eep neural networks'} ability to model complex patterns and relationships in data makes them susceptible to overfitting. Hence, regularization is essential to ensure generalization and prevent memorization of the training data. Conventional regularization techniques, such as Dropout, Gaussian Dropout~\cite{srivastava2014dropout}, and DropConnect~\cite{wan2013regularization}, incorporate stochasticity by randomly deactivating the weights associated with the neurons during training.

Dropout typically applies a Bernoulli random mask per input, reducing neuron outputs probabilistically, whereas Gaussian Dropout multiplies neuron outputs with Gaussian noise. Both approaches introduce randomness into the neural network activations on a per-input basis while maintaining a consistent perturbation pattern across all nodes within a layer. DropConnect applies a Bernoulli mask to weights, which remains uniform across batches. Ensemble-inspired methodologies, such as Masksembles~\cite{durasov2021masksembles}, replicate diverse model behaviors by applying specified binary masks to subnetworks, thereby enabling ensemble-like inference within a single model. However, these masks remain static during training, limiting the stochasticity of regularization. The proposed PerNodeDrop approach breaks this homogeneity  by incorporating per-sample, per-connection stochasticity using Bernoulli or Gaussian noise, creating a nuanced regularization layer.

PerNodeDrop is a per-sample, per-node regularization layer that introduces fine-grained, structured stochastic perturbations, offering greater diversity than prior work. It can operate in either binary or Gaussian modes, with fixed or dynamic masks, providing a user-configurable stochastic behavior that is compatible with common hyperparameter search APIs. This design produces an implicit ensemble effect at low computational cost by leveraging the dynamic subnetworks explored during training. The method was evaluated across image, text, and audio tasks, and showed stable training, reduced validation loss, and indications of improved resistance to overfitting.

\section{Related Work}
Overfitting is common in deep neural networks, prompting the study and development of approaches that add a regularization layer to the network composition. These methods induce stochasticity at multiple levels, including neurons, connections, and structural features, by using mechanisms such as dropout or Gaussian perturbations. The applicability of certain strategies is uniform across an entire layer of nodes, or across a batch of inputs, resulting in consistent noise being applied to the target layer or the target batch of inputs.

The theoretical basis for noise-based regularization was established by Bishop ~\cite{bishop1995training}, who showed that training with additive noise is mathematically equivalent to Tikhonov regularization, thereby justifying stochastic perturbation methods.An ~\cite{An1996} and Neelakantan et al.~\cite{Neelakantan2015} showed that stochasticity can smooth the optimization landscape, aiding convergence and generalization.

\subsection{Dropout and Gaussian Dropout}

Stochastic regularization can be represented as a multiplicative operation on the input:
\begin{equation}
y = m \cdot x
\end{equation}
where $x$ is the input signal, $y$ is the output signal, and $m$ is an element-wise stochastic mask applied. The formulation of $m$ varies depending on the regularization method. For example, in \textbf{Dropout}~\cite{srivastava2014dropout} the most widely adopted regularization technique applies a Bernoulli random mask to the input, where each element of $x$ is independently set to zero with probability $p$ and retained with probability $1-p$. Formally:
\begin{equation}
m \sim \text{Bernoulli}(1 - p), \quad y = m \cdot x
\end{equation}

Wager et al.~\cite{Wager2013} conceptualized dropout as a method of adaptive regularization within a generalized linear model framework, providing theoretical understanding of its data-dependent penalty impact.

\textbf{Gaussian Dropout}~\cite{srivastava2014dropout}  is a closely related variant of the standard dropout. Instead of applying a binary (Bernoulli) mask, it applies multiplicative Gaussian noise, resulting in a seamless and continuous form of stochasticity. The Gaussian Dropout multiplies the input with noise sampled from a normal distribution, i.e., $m \sim \mathcal{N}(1, \sigma^2)$. The resulting output is given by:

\begin{equation}
m \sim \mathcal{N}(1, \sigma^2), \quad y = m \cdot x
\end{equation}

\subsection{DropConnect}

\textbf{DropConnect}~\cite{wan2013regularization} introduce dropout at the weight level. For a weight matrix $W$ and mask $M \sim \text{Bernoulli}(1 - p)$, the effective weight becomes $W^* = W \odot M$, where $\odot$ denotes element-wise multiplication. The forward pass is:
\begin{equation}
y = x \cdot (W \odot M)
\end{equation}
However, this method often uses a single mask per mini-batch, but if applied to individual inputs, it may result in better generalization.

\subsection{Ensemble Methods and Mask-Based Routing}
Ensemble approaches provide robust generalization and uncertainty estimates by consolidating the predictions from numerous independently trained neural networks. Although Baldi and Sadowski ~\cite{Baldi2014} interpreted dropout as an implicit ensemble method and acknowledged the potential loss of advantageous co-adaptations, efforts have been made to implement explicit ensembles to further improve the model. Nevertheless, conventional \textbf{Deep Ensembles} ~\cite{Lakshminarayanan2017} entail significant computational and memory expenses owing to the requirement of training and retaining many complete models.
Addressing this, \textbf{Snapshot Ensembles}~\cite{Huang2017Snapshot} advocated preserving intermediate checkpoints from a singular training trajectory with a cyclical learning rate, thus behaving as an ensemble without supplementary training. \textbf{Masksembles}~\cite{durasov2021masksembles} enhanced the notion of efficient ensembling by simulating the ensemble behavior within a single model by activating designated subnetworks using predetermined binary masks. 
Masksembles use fixed binary masks on neuron clusters, requiring group sizes that evenly divide a layer’s input dimensions and preserve an identical masking pattern throughout training. This reduces stochastic variability and regularization relative to approaches that modify masks per input or batch. While Masksembles can approximate ensemble behavior at a lower cost, their static routing limits the model’s ability to explore representational diversity during training,  as seen in the recent empirical reassessments of Masksembles ~\cite{MasksemblesReeval2024}  under standard benchmarking protocols.  \textbf{BranchyNet}~\cite{Teerapat2016} introduces early-exit branches that halt processing once a high-confidence prediction is obtained, improving inference efficiency but yielding benefits that depend on input confidence.

\subsection{Adaptive Routing-Based Structures}

Routing-based methods perform input-conditioned path selection, enabling specialized feature processing rather than a static feedforward computation flow.. Unlike BranchyNet, where routing primarily enables early termination, the routing mechanisms considered here promote feature routing and expert selection.

Dynamic routing among capsules, introduced in \cite{Sabour2017}, iteratively updates coupling coefficients according to an agreement between lower- and higher-level capsules, encouraging structured representations that preserve part–whole and spatial relationships. While this yields interpretable, hierarchical features, the routing iterations are computationally costly and can be sensitive to noise and initialization.
 
Mixture-of-Experts (MoE) models \cite{Shazeer2017} use a learned gating network to activate only a small subset of expert subnetworks per input. This enables high model capacity with limited additional computation, but stable training requires effective load-balancing to prevent expert collapse and routing instability.

Routing-based mixture-of-experts models provide input-dependent specialization but often suffer from expert collapse, sensitivity to routing hyperparameters, and domain-specific routing procedures. Many routing mechanisms are only partially differentiable or rely on additional architectural components, which can limit robustness and portability across modalities.

PerNodeDrop achieves a related form of input-dependent variation without explicit routing or discrete selection. Neurons are probabilistically activated or suppressed on a per-sample basis, allowing different subnetworks to be explored during training because the stochastic masks are differentiable in expectation. This avoids expert collapse and does not require gating, dispatching, or domain-specific routing logic. As a result, PerNodeDrop provides a lightweight alternative to MoE architectures: it introduces controlled specialization and input-adaptive behavior with none of the computational or structural overhead associated with routing-based models.

\subsection{Bayesian and Noise-Based Extensions}

Gal and Ghahramani~\cite{Gal2016} showed that applying dropout at test time and averaging multiple stochastic forward passes approximates a Bayesian posterior over models as follows:
\begin{equation}
\mathbb{E}[y] = \frac{1}{T} \sum_{t=1}^{T} f(x; m^{(t)}), \quad m^{(t)} \sim \text{Bernoulli}(1 - p)
\end{equation}
This links dropout to approximate inference in deep Gaussian processes. 

\textbf{Monte Carlo (MC) Dropout}~\cite{Gal2016}, leverages a Bayesian interpretation of standard dropout by approximating model uncertainty through stochastic sampling at inference time, instead of deactivating dropout or noising during inference as implemented in most cases. MC Dropout uses dropout to make many forward passes and sample the posterior distribution of the models. This allows the approximation of Bayesian inference in deep neural networks without ensembles or variational inference techniques. Despite the numerous stochastic passes adding inference-time overhead, this strategy is intriguing due to its simplicity and compatibility with regular training methods.

\textbf{Variational Dropout}~\cite{Kingma2015} learns dropout rates by minimizing the KL divergence between approximate and true posteriors. Variational Dropout relies on variational inference, which complicates parameter sampling and training objectives. Its performance depends strongly on the choice of prior and learning rate, and large noise levels can produce high-variance gradients that destabilize training and lead to suboptimal convergence.

\textbf{R-Drop}~\cite{Liang2021} introduced a symmetric KL loss between two stochastic forward passes:
\begin{equation}
\mathcal{L}_{\text{R-Drop}} = \text{KL}(P_1 \parallel P_2) + \text{KL}(P_2 \parallel P_1)
\end{equation}
This encourages output consistency but incurs additional computational costs owing to the need for dual forward passes per input. Furthermore, reliance on the proximity of the two stochastic predictions may implicitly enforce redundancy in internal representations.

\textbf{Markov Dropout}~\cite{Korzepa2019Markov} Rooted in the Bayesian interpretation of dropout as an approximate inference, introduces temporal correlation into the dropout process by modeling the dropout mask evolution as a first-order Markov chain, designed for recurrent networks and other sequential modeling, where preserving temporal coherence in hidden state dynamics is critical. The applicability does not naturally extend to the conventional feedforward or convolutional networks where the temporal axis is irrelevant.

\subsection{Other Extensions}
\textbf{Stochastic Depth}~\cite{Huang2016StochasticDepth} skips entire layers with a probability $p$ during training. Similarly, Scheduled \textbf{DropPath}~\cite{Larsson2017FractalNet} adjusts the drop probability across epochs.During training, \textbf{Stochastic Depth}{\cite{Huang2016StochasticDepth} skips complete layers with probability $p$. Similarly Scheduled \textbf{DropPath}~\cite{Larsson2017FractalNet} adjusts drop probability over time.Extensions such as \textbf{SpatialDropout}{\cite{Tompson2015SpatialDropout} and \textbf{DropBlock}~\cite{Ghiasi2018DropBlock} improve dropout mask spatial correlation in convolutional networks, but do not address input-specific stochasticity.Their stochastic techniques are less flexible, hence they may not generalize well across diverse architectures or input domains without substantial tunings.

\subsection{Co-Adaptation, Redundancy, and Specialization}

~\cite{Hinton2012Dropout} proposed dropout to decrease co-adaptation:- however excessive randomness may result in redundant features, hence under-capacitating the model ~\cite{Omathil2025Redundancy}.Arpit et al.~\cite{Arpit2017Memorization} demonstrate that over-regularized deep networks memorize data, thereby emphasizing the need for regulated diversity.

In ensemble approaches such as Masksembles~\cite{durasov2021masksembles},  subnetworks can specialize to attend to diverse features of data but may lack dynamic behavior, whereas routing-based solutions add structure but complicate domain generalization.

\subsection{Survey studies and trends in recent developments}
Li et al.\cite{Li2021Survey} examined deep-learning regularization and optimization strategies, highlighting recent trends that mitigate the limitations of dropout-based methods by mechanism and effect.

\textbf{Structured noise injection.} Injects noise at selected layers or connections to improve robustness while preserving semantic structure. Examples include DropConnect~\cite{wan2013regularization}, which randomly removes weights during training, and Shakeout~\cite{Kang2019Shakeout}, which applies input-dependent noise to scale activations and induces composite regularization effects (e.g., $L_{0}$ sparsity, $L_{1}$ shrinkage, and $L_{2}$ weight decay).

\textbf{Input-aware regularization.} Modulates regularization according to input attributes to preserve task-relevant representations. Variational Dropout (Kingma, 2015) estimates dropout rates via probabilistic inference, whereas Adaptive Dropout (Ba, 2013) alters dropout probabilities conditioned on input variables.

\textbf{Dynamic dropout strategies.} Vary dropout schedules during training to improve convergence and feature learning. Examples include Curriculum Dropout\cite{Morerio2017CurriculumDropout}, which progressively increases dropout, and Annealed Dropout\cite{Pham2019AnnealedDropout}, which gradually decreases dropout for later-stage stability.

\textbf{Scalable and context-sensitive methods.} Recent work emphasizes context-aware mechanisms: Noise Injection Node Regularization (NINR)\cite{Levi2022NINR} uses auxiliary nodes to inject structured noise for domain-shift robustness; Adaptive Tabu Dropout\cite{Kumari2021TabuDropout} prevents recently dropped neurons from immediate reactivation to enhance variability; and Gradient Routing\cite{Cloud2024} applies input-sensitive masks to guide subnetwork learning.Recent work increasingly explores architectural flexibility and fine-grained, input-dependent stochasticity. PerNodeDrop extends this idea by applying per-sample, per-connection masking, unifying concepts from several existing regularization methods.

\subsection{Identified gaps and motivation for PerNodeDrop}
Despite the breadth of existing regularization methods, several limitations remain:
\begin{itemize}
\item \textbf{Limited input-dependent stochasticity:} Many methods use identical masks across layers or batches, reducing per-sample variability.
\item \textbf{Restricted configurability:} Most methods support only fixed noise types and do not allow flexible selection of binary, Gaussian, or user-defined perturbations, nor easy switching between fixed and dynamic masking schemes.

\item \textbf{Insufficient granularity:} Per-sample, per-connection control is uncommon.
\item \textbf{Limited compatibility with automated search:} Many regularizers rely on manual hyperparameter tuning and are not designed to integrate naturally with Bayesian or evolutionary optimization frameworks.
\end{itemize}

PerNodeDrop addresses these gaps by:
\begin{itemize}
\item Applying noise at \textbf{per-sample, per-connection} granularity;
\item Supporting multiple noise models (e.g., \textbf{Gaussian} and \textbf{binary}) and user-defined functions;
\item Allowing \textbf{fixed or dynamic masks} depending on the training regime;
\item Providing \textbf{tunable architecture parameters} to enable automated hyperparameter optimization.
\end{itemize}

Building on earlier work that saw dropout as approximate Bayesian inference using Monte Carlo sampling~\cite{Gal2016},Kendall and Gal~\cite{Kendall2017} distinguished between epistemic and aleatoric uncertainty, emphasizing the influence of stochastic processes on model confidence as opposed to deterministic prediction. From this broader perspective, PerNodeDrop may be interpreted as a structured, per-sample stochastic masking mechanism that adapts across both individual instances and network connections. Core idea is to discourage unnecessary pathways while encouraging important predictive routes during training, which matches the ideas behind the Information Bottleneck~\cite{Tishby1999}, helping the model learn representations that are both simple and informative.
This targeted noise injection enhances activation diversity across forward passes, thereby expanding the explored function space without the need for explicit ensembles.

\section{Background}
Deep neural networks can capture hierarchical structure in data but are susceptible to overfitting, as they frequently exploit dataset-specific associations, which may pertain in training data set; hence, this is often termed "memorization vs. generalization". Arpit et al.~\cite{Arpit2017Memorization} demonstrated that deep models can memorize arbitrary labels, emphasizing the need for robust regularization.

\textit{Co-adaptation} among neurons happens when units become too dependent on the activations of others, reducing generalization and increasing the risk of overfitting~\cite{Hinton2012Dropout}. Stochastic regularization methods such as Dropout~\cite{srivastava2014dropout}, DropConnect~\cite{wan2013regularization}, and Shakeout~\cite{Kang2019Shakeout} compat this by injecting noise into the network during training. Dropout applies a probabilistic mask to neuron activations at each forward pass, while DropConnect randomly masks weights and therefore modifies the connectivity pattern. Although DropConnect removes the layer-wise uniformity of Dropout, it typically applies a single weight mask across a mini-batch, introducing batch-level uniformity.

Bengio et al.\cite{Bengio2013} indicated that some degree of co-adaptation or neuron specialization may actually be beneficial; this was experimentally supported by Omathil et al. ~\cite{Omathil2025Redundancy} with a claim that regularization could form redundancy, causing delayed convergence of transfer as it breaks the diverse specialization of neurons. Ensemble-based methods~\cite{Lakshminarayanan2017} promote subnetwork specialization but require costly retraining of multiple models. Snapshot Ensembles~\cite{Huang2017Snapshot} mitigate this computational burden by leveraging the intermediate checkpoints of a single training run to create an ensemble, thereby avoiding the need to train models independently. Snapshot Ensembles provide diverse function approximators, but a true subnetwork specialization cannot be expected, as it lacks network partitioning.

Gal and Ghahramani~\cite{Gal2016} showed that Monte Carlo Dropout at test time can be interpreted as an approximate Bayesian inference technique, allowing models to estimate epistemic uncertainty in their parameters. Kendall and Gal~\cite{Kendall2017} later distinguished between \textit{epistemic} and \textit{aleatoric} uncertainty, arguing that both are essential for reliable predictions. They proposed probabilistic loss functions, such as heteroscedastic negative log-likelihood, which enable models to learn input-dependent noise and capture aleatoric uncertainty.

Although probabilistic views prompted uncertainty-aware regularizations such as Variational Dropout~\cite{Kingma2015}  and R-Drop~\cite{Liang2021}, Variational Dropout requires hyperparameter adjustment and reparameterization, whereas R-Drop adds computing complexity from symmetric KL losses. Structured techniques such as Markov Dropout~\cite{Korzepa2019Markov} and Stochastic Depth~\cite{Huang2016StochasticDepth}  increase temporal or layer-level coherence lacking fine-grained connection-level management. Similarly, spatial techniques like DropBlock~\cite{Ghiasi2018DropBlock} introduce structured noise but are spatially constrained.

PerNodeDrop contributes to this line of work by introducing a per-sample, per-connection noise mechanism that supports both binary and Gaussian perturbations in fixed and dynamic modes. By varying these masks across nodes and training iterations, the network is encouraged to adapt its weights under more diverse perturbation patterns.

Its design aligns with the Information Bottleneck principle~\cite{Tishby1999}, promoting compact yet informative latent representations by removing irrelevant pathways. In doing so, it enables controlled diversity without sacrificing learning stability.

\subsection{Design Criteria for Modern Regularization}
To address the trade-off between specialization and generalization, a robust regularization framework could be:
\begin{itemize}
    \item Capture input-specific uncertainty using per-sample stochasticity;
    \item Permit per-connection noise masking for fine-grained control;
    \item Support both binary and Gaussian noise types;
    \item Enable fixed or dynamic masking policies;
    \item Remain compatible with tuning frameworks like Bayesian optimization~\cite{Snoek2012}.
\end{itemize}
PerNodeDrop was built with these principles, unifying several trends in stochastic regularization into a single extensible framework with configurable hyperparameters.

\section{Proposed Methodology}

This work proposes \textbf{PerNodeDrop}, a fine-grained regularization framework that applies stochastic masks at a localized, per-sample, per-connection level, applying different masks to incoming signals for different nodes of the layer, allowing a unique subnetwork to be activated for each sample.

PerNodeDrop is governed by two key hyperparameters: 
\emph{Stir Type}, which defines the perturbation function employed for stochastic regularization, 
and the \emph{Drop Rate}, which controls the magnitude of perturbation and the \emph{Dense Units Count/Activation Function}, which defines the dense node count and the activation function such as RELU, where each node is assigned an independent randomization as per the \emph{Stir Type} and \emph{Drop Rate}.
Together, these parameters determine the stochastic behavior and strength of the regularization effect.

\subsection{Comparison with DropConnect}

DropConnect~\cite{wan2013regularization} regularizes networks by applying a Bernoulli mask directly to the weight matrix during training:
\[
z^{(k)} = (W \odot M) x^{(k)}, \quad M \sim \text{Bernoulli}(1 - p)
\]
where \(M\) is sampled once per forward call and shared across all samples in the mini-batch.  
This design introduces \emph{batch-level uniformity}, meaning that all the inputs in the batch experience the same sparsified connectivity pattern.

In contrast, the proposed  \textit{PerNodeDrop} layer generates an independent mask for each input example:
\[
z^{(k)} = (W \odot M^{(k)}) x^{(k)}, \quad M^{(k)} \sim \text{Bernoulli}(1 - p),
\]
where \(M^{(k)} \in \mathbb{R}^{D_{\text{in}}\times D_{\text{out}}}\) is a binary or Gaussian mask unique to the \(k\)-th input, producing sample-level subnetworks that encourage functional diversity.  
Figure~\ref{fig:PerNodeDrop_comparison} schematically contrasts both formulations.

\textbf{Mathematical Equivalence and Distinction:}

Both DropConnect and PerNodeDrop perturb the matrix product $C = AB$, differing in which term is stochastically masked: 
DropConnect applies a binary mask $M_B$ to the weights $B$, $C = A(B \odot M_B)$, 
whereas PerNodeDrop applies a mask $M_A$ to the incoming activations $A$, $C = (A \odot M_A)B$, while the expected output of both satisfies \(\mathbb{E}[C] = (1-p)AB\), the variance structure and computational behavior differ.

DropConnect introduces localized edge-level sparsity by dropping individual weights, whereas PerNodeDrop applies structured signal-level suppression that remains fully compatible with dense matrix kernels, thereby improving computational efficiency.

\textbf{Implementation Challenges in DropConnect:}
As noted by Wan \textit{et al.}~\cite{wan2013regularization}, per-element masking causes substantial memory and access overheads:
\begin{quote}
``Since the memory requirement for the \(M\)'s now grows with the size of each mini-batch, the implementation needs to be carefully designed as described in Section~5.''
\end{quote}
They addressed this by (i) storing each mask element as a single bit, reducing memory by \(32\times\), and (ii) using 2D texture-aligned GPU memory to improve access bandwidth.  
These optimizations enabled efficient per-example DropConnect but are rarely reproduced in general-purpose frameworks owing to implementation complexity.

\textbf{Practical Implementations:}
Most modern libraries (PyTorch, TensorFlow) adopt a \emph{mini-batch–shared} variant for efficiency.  
For example, the PyTorch-NLP implementation wraps layers using a \texttt{WeightDrop} function:

\lstdefinestyle{IEEEstyle}{
  basicstyle=\ttfamily\footnotesize,
  numbers=none,
  breaklines=true,
  frame=single,
  backgroundcolor=\color{gray!5},
  captionpos=b,
  aboveskip=3pt,
  belowskip=3pt,
  showstringspaces=false
}
\lstset{style=IEEEstyle}

\begin{lstlisting}[language=Python,
caption={PyTorch practical DropConnect (WeightDrop) implementation},
label={lst:pytorch_dropconnect},
breaklines=true]
def forward(*args, **kwargs):
    for name_w in weights:
        raw_w = getattr(module, name_w + '_raw')
        w = F.dropout(raw_w, p=dropout, training=module.training)
        setattr(module, name_w, w)
    return original_module_forward(*args, **kwargs)
\end{lstlisting}

Here, \texttt{F.dropout} applies a single Bernoulli mask to the entire weight tensor per forward pass.
Because this mask has no batch dimension, all samples in the mini-batch share the same masked weights—confirming batch-level DropConnect behavior.

\textbf{PerNodeDrop Implementation:}
In contrast, PerNodeDrop applies stochastic masks directly to input activations using tensors that explicitly include the batch dimension.

\begin{lstlisting}[language=Python,
caption={PerNodeDrop (PerNodeDrop) core implementation},
label={lst:pernode_core},
breaklines=true]
# Mask applied per input (includes batch dimension)
mask = tf.nn.dropout(tf.ones_like(inputs), rate=drop_rate)
inputs_masked = inputs * mask
output = tf.matmul(inputs_masked, W)
\end{lstlisting}

Because the mask shape is \((\text{batch}, D_{\text{in}})\) or \((\text{batch}, D_{\text{in}}, D_{\text{out}})\),
each sample receives an independent stochastic masking pattern.

\textbf{Computational Observations:}
On the CIFAR-10 and RCV datasets, PerNodeDrop exhibited runtime broadly comparable to standard Dropout.
However, for higher-dimensional inputs such as spectrograms of size \(128\times128\times1\),
the computational overhead increases, but sublinearly with input size owing to efficient vectorized operations
and GPU-friendly dense kernel fusion. The detailed timing results are provided in TABLE IV

\begin{figure}[t]
  \centering
  \includegraphics[width=0.95\linewidth]{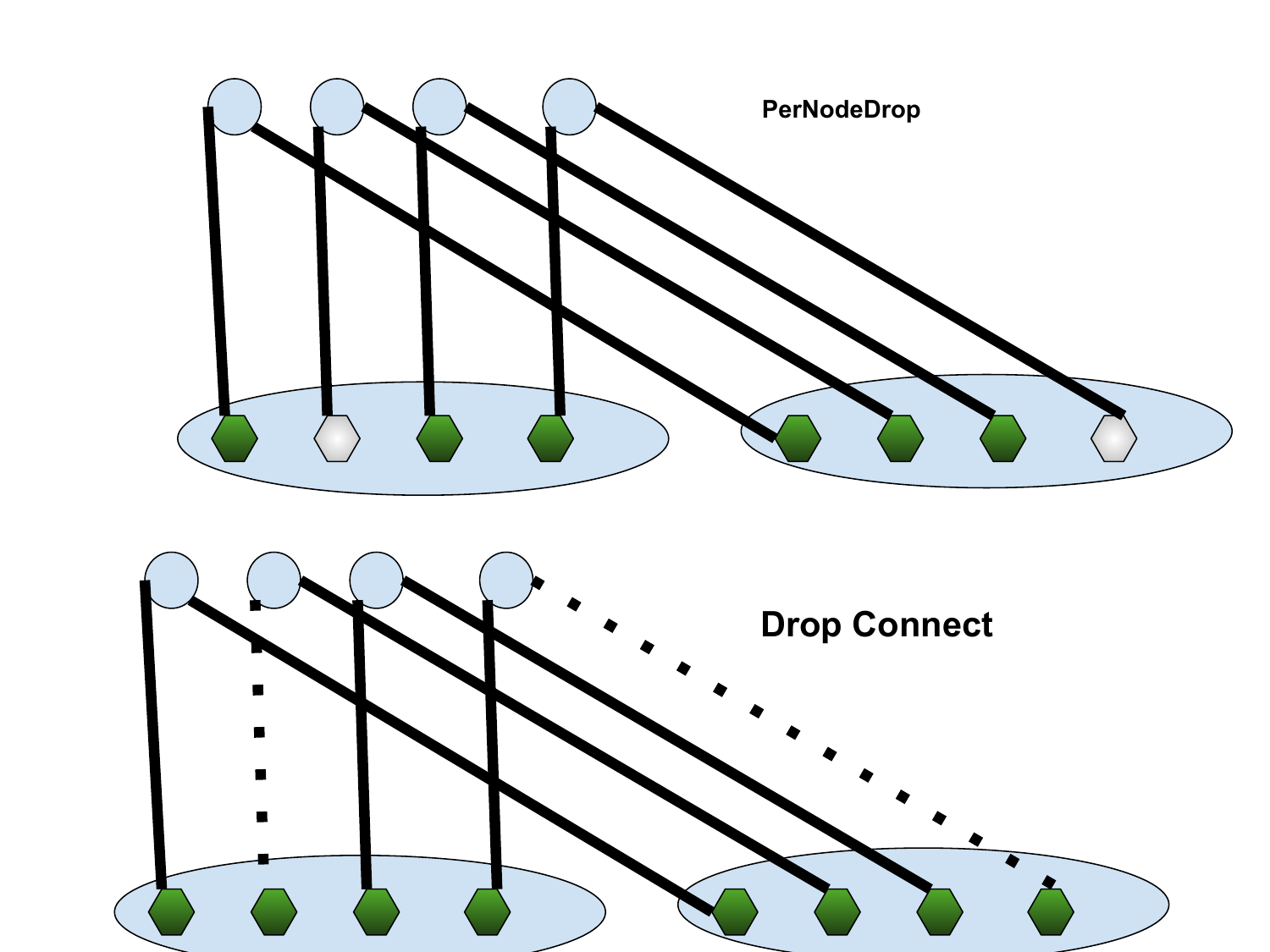}
  \caption{Comparison of DropConnect and PerNodeDrop.
  (Top) DropConnect applies a single mask to weights shared across the mini-batch,
  yielding one subnetwork per batch.
  (Bottom) PerNodeDrop assigns each input its own mask,
  producing per-sample subnetworks with distinct signal paths.}
  \label{fig:PerNodeDrop_comparison}
\end{figure}

\textbf{Efficient Implementation:}
Despite the fine-grained granularity in noise injuction, PerNodeDrop is computationally efficient due to the use of:
\begin{itemize}
    \item \textbf{Batch-wise vectorized mask sampling}: Masks are sampled in a single tensor operation with shape \((B, D_{\text{in}}, D_{\text{out}})\), where \(B\) is batch size.
    \item \textbf{Broadcasted einsum operations}: Forward passes are computed using optimized einsum kernels that avoid loops while respecting sample-wise differences.
    \item \textbf{No per-sample kernel duplication}: Weight parameters are shared; only the effective connection paths vary per input.
\end{itemize}

This approach avoids the inefficiency of DropConnect with batch size = 1 (which would simulate similar behavior but at high computational cost).

\subsection{Gaussian Stir Type}
PerNodeDrop also supports a Gaussian noise variant, inspired by the Gaussian Dropout~\cite{wang2013fast}. Instead of binary masking, the connection weights were scaled using normally distributed noise:
\[
M^{(k)}_{i,j} \sim \mathcal{N}(1, \sigma^2), \quad \tilde{W}^{(k)} = W \odot M^{(k)}
\]
This injects multiplicative noise to each weight per sample, encouraging smoother regularization and variance-aware activation patterns.

The stir type could also be extended for example a 'partial-Gaussian variant'  where a subset of connections receive Gaussian noise 
while others remain unperturbed:
\[
M^{(k)}_{i,j} =
\begin{cases}
1, & r_{i,j}^{(k)} > p,\\[4pt]
\mathcal{N}(1, \sigma^2), & \text{otherwise,}
\end{cases}
\]
where \(r_{i,j}^{(k)} \sim \mathcal{U}(0,1)\) is a uniform random variable controlling  which connections are perturbed, by comparing to a configurable threshold value p(0-1). This mechanism enables mixed stochasticity, combining deterministic and Gaussian-masked weights to balance the stability and noise diversity.

\subsection{Fixed vs. Dynamic Masking}
\label{sec:masking}

\emph{PerNodeDrop} supports two masking regimes:
\begin{itemize}
\item \textbf{Dynamic.} A new mask was sampled at each forward pass, producing distinct subnetworks across iterations and thereby providing stochastic regularization.
\item \textbf{Fixed.} A mask is sampled once and retained for the duration of training, effectively defining a persistent sparse subnetwork for each input or model instance.
\end{itemize}

In contrast to \emph{Masksembles}~\cite{durasov2021masksembles}, which groups neurons and requires the group size to divide the input dimension of a layer, \emph{PerNodeDrop} imposes no such constraints. Neurons are masked independently, allowing flexible and irregular sparsity patterns. During evaluation, dynamic-mask variants disable stochastic sampling and scale activations by the expected mask value, consistent with standard dropout inference. Fixed-mask variants retain their learned sparse subnetworks, which act as architectural priors.

\subsection{Expected-Loss Interpretation}

Noise-based regularization is commonly analyzed using a second–order Taylor
expansion of the loss under a multiplicative mask. Let $\mathbf{W}$ denote the
weights and $\mathbf{M}$ a random elementwise mask. With the centered
perturbation $\Delta\mathbf{M}=\mathbf{M}-\mathbb{E}[\mathbf{M}]$, we write
\begin{equation}
\mathbf{W}\odot\mathbf{M}
=
\mathbf{W}\odot\mathbb{E}[\mathbf{M}]
+
\mathbf{W}\odot\Delta\mathbf{M},
\label{eq:mask_decomp_new}
\end{equation}
and expand $L(f(x;\mathbf{W}\odot\mathbf{M}))$ around $\mathbf{W}$:
\begin{align}
L(f(x;\mathbf{W}\odot\mathbf{M}))
&\approx
L(f(x;\mathbf{W}))
+
\langle\nabla_{\mathbf{W}}L,\,
\mathbf{W}\odot\Delta\mathbf{M}\rangle
\nonumber\\[-3pt]
&\quad+
\tfrac{1}{2}
(\mathbf{W}\odot\Delta\mathbf{M})^{\top}
\mathbf{H}_{L}
(\mathbf{W}\odot\Delta\mathbf{M}),
\label{eq:taylor_expand_new}
\end{align}
where $\mathbf{H}_{L}$ is the Hessian of $L$ with respect to $\mathbf{W}$.
Taking expectations cancels the linear term.

\subsubsection{Bernoulli Dropout (Wager \emph{et al.})}
For masks $M_i\!\in\!\{0,1\}$ with
$\mathbb{E}[M_i]=1-p$ and $\operatorname{Var}(M_i)=p(1-p)$,
Wager \emph{et al.}~\cite{Wager2013} obtained, for generalized linear models,
\begin{equation}
\mathbb{E}[L(f(x;\mathbf{W}\odot\mathbf{M}))] 
\approx 
L(f(x;\mathbf{W}))
+
\tfrac{1}{2}\sum_{i}
p(1-p)\,W_i^{2}\,
\frac{\partial^{2} L}{\partial W_i^{2}}.
\label{eq:wager_equation_new}
\end{equation}

\subsubsection{Gaussian Multiplicative Noise}
For mean–preserving Gaussian masks $M_i\!\sim\!\mathcal{N}(1,\sigma^{2})$,
$\Delta M_i$ has variance $\sigma^{2}$.  
Substituting into \eqref{eq:taylor_expand_new} gives
\begin{equation}
\mathbb{E}[L(f(x;\mathbf{W}\odot\mathbf{M}))]
\approx
L(f(x;\mathbf{W}))
+
\tfrac{1}{2}\sum_{i}
\sigma^{2}W_i^{2}
\frac{\partial^{2}L}{\partial W_i^{2}},
\label{eq:gaussian_equation_new}
\end{equation}
which reduces to Bishop’s classical noise penalty\footnotemark[1] 
under a locally isotropic Hessian.

\footnotetext[1]{Bishop~\cite{bishop1995training} analyzed additive Gaussian input noise; Wager \emph{et al.}~\cite{Wager2013} provided an analogous result for Bernoulli dropout.}

\subsubsection{Interpretation for PerNodeDrop}
PerNodeDrop fits into the same framework but uses finer per–neuron and
per–sample masks.  
Let $\Sigma_{\Delta M}=\operatorname{Cov}(\Delta\mathbf{M})$ denote the
covariance of the mask perturbation.  
Taking expectations of \eqref{eq:taylor_expand_new} yields the general form
\begin{equation}
\mathbb{E}[L]
\approx
L
+
\tfrac{1}{2}
\operatorname{tr}\!\left(
\mathbf{H}_{L}\,
\operatorname{diag}(\mathbf{W})\,
\Sigma_{\Delta M}\,
\operatorname{diag}(\mathbf{W})
\right),
\label{eq:pernode_general_new}
\end{equation}
which reduces to the diagonal expressions in 
\eqref{eq:gaussian_equation_new} when mask components are independent.

Independent Gaussian-like masks recover the standard variance–weighted
penalty.  
However, PerNodeDrop’s per-neuron construction introduces heterogeneous
variances (and possible weak correlations) across units because masks are
drawn independently for each neuron and sample.  
This leads to a curvature penalty acting along more directions of
$\mathbf{H}_{L}$ than the uniform-diagonal noise used in conventional dropout,
and results in smoother optimization and reduced overfitting.

\textbf{Intuition}
Let $f(x;\theta)$ denote a neural network, and let $\mathcal{N}$ denote a stochastic perturbation operator. Conventional methods average noise effects across training iterations. In contrast, PerNodeDrop applies independent perturbations $\epsilon_i$ to each unit, with $\epsilon_i \perp\!\!\!\perp \epsilon_j$ for $i \neq j$, enabling partial intra-sample averaging of zero-mean noise through activation aggregation. Hence the independent masks supply small localized perturbations to neuronal activations. Each forward pass effectively follows a slightly different functional pathway, offering lightweight model diversity without extra networks. This reduces stochastic variance while preserving consistent signal structure, preventing reliance on spurious feature–label coincidences and narrowing training–validation discrepancies.

\subsection{Summary}

PerNodeDrop offers a tunable and extensible regularization framework with the following configuration axes:
\begin{itemize}
    \item \textbf{Stir Type}: A hyperparameter that determines the masking distribution. It can be set to \emph{Binary} (Dropout-style), \emph{Gaussian}, or a \emph{hybrid partial-Gaussian} mode, and can be extended to other continuous perturbation functions.
    \item \textbf{Masking Mode}: Fixed or Dynamic
    \item \textbf{Granularity}: Per-connection, per-sample masking
\end{itemize}

This formulation generalizes and extends multiple existing techniques:
\begin{itemize}
    \item \textbf{Dropout} is recovered as dynamic binary masking at the neuron level.
    \item \textbf{DropConnect} is a special case with batch-shared masks.
    \item \textbf{Masksembles} corresponds to fixed binary masking.
    \item \textbf{Gaussian Dropout} is extended to per-connection Gaussian perturbations.
\end{itemize}

\section{Experimental Setup}

A modular deep learning experimentation framework was implemented to evaluate stochastic regularization methods across multiple data modalities, including image classification, natural language processing, and signal processing. The framework supports dynamic insertion of regularization layers, enabling fair and reproducible comparisons among model variants. Each experiment records detailed training logs for post hoc analysis. Preprocessing of data specific to the data domain was done, which includes image normalization for vision tasks and tokenization for text-based tasks.

All the experiments used a common backbone appropriate to the data domain (e.g., convolutional networks for image tasks). The architecture provides interchangeable slots for different regularization modules, followed by a fully connected \texttt{Dense}  layer. Hyperparameters specify both the regularization method and the Dense-layer width (\texttt{units}).

For PerNodeDrop, the regularization module is composite and contains an internal Dense transformation in which each neuron is paired with an independent masking function. To ensure fair comparison, all other configurations included an additional Dense layer of identical width appended after their regularization module.

\subsection{Training Configuration}

Each model is trained using a standardized configuration:
\begin{itemize}
    \item \textbf{Dropout Rates:} \{0.0, 0.1, \dots, 0.9\}
    \item \textbf{Dense Layer Widths:} e.g., 32, 64, 128
    \item \textbf{Batch Sizes:} e.g., 64, 128, 256
    \item \textbf{Epochs:} Typically 20 (with optional early stopping)
    \item \textbf{Optimizer:} Adam
    \item \textbf{Loss Function:} Categorical crossentropy (or task-specific)
\end{itemize}

\subsection{Evaluation Metrics and Logging}

Two key metrics are tracked at the end of each epoch:
\begin{enumerate}
    \item \textbf{Validation Loss (\texttt{val\_loss}):} Used for model selection and generalization assessment.
    \item \textbf{Epoch-End Training Loss:} Computed by evaluating the fully updated model on the entire training dataset after the epoch was completed.
\end{enumerate}

As the Keras API's train loss is influenced by intermittent parameter adjustments and stochastic regularization effects, making it unsuitable for direct comparison with validation metrics. To compare the model's true performance on trainin data, after all parameter updates within an epoch, the \textbf{epoch-end training loss} was recorded using a custom callback. These metrics and dropout rate were logged for each training setting. For each regularization variant, the top three configurations (based on least validation loss) across all dropout rates were chosen and their epochs and training–validation losses were presented for comparison.

For each regularization variant, the validation and epoch-end training losses were logged along with their associated hyperparameters (notably, the dropout rate). 
From the full training history logs across all variants and hyperparameter configurations, the lowest validation-loss values were identified across dropout rates for each variant.These distinct best-per-rate points were then visualized in two complementary ways:
\begin{itemize}
    \item \textbf{Bar plots:} The lowest three validation losses per variant were plotted to provide a direct comparison of the best-performing configurations across regularization types, sorted on the basis of validation loss from lowest to  highest.
    \item \textbf{Scatter plots:} Generalization trends were illustrated by graphing validation loss vs epoch-end training loss for the three optimal validation losses per variant recorded during training. Configurations with low training loss and elevated validation loss may indicate overfitting; however while validation losses are very proximal, a relatively larger training loss may suggest superior generalization. 
\end{itemize}

A non-parametric statistical method was used to rank and validate the relative performance of the regularization alternatives, where each variant, the five best records were selected based on the lowest validation losses, ensuring an equitable comparison of each technique under the corresponding most favorable conditions. In this selection, duplicate dropout rates were permitted, because the objective was to identify optimal performance for each variant. These selected validation losses were ranked, and the rank distributions were used to conduct a Friedman test, a non-parametric analogue of repeated-measures ANOVA, to assess whether the observed performance differences among the regularization variants were statistically significant. The test computes a Friedman statistic $(\chi^{2})$ with an associated $p$-value, along with Kendall’s coefficient of concordance $(W)$, which measures the consistency of rankings across repeated trials. This ranking-based methodology offers a statistical basis for comparing regularization strategies across varied configurations of architecture and hyperparameters

\section{Experimental Results and Analysis}

\subsection{Image Domain: CIFAR-10}

The \textbf{CIFAR-10 dataset} was utilized and normalized to the range [0,1], with labels converted to one-hot encoding. 
The CNN architecture comprises two convolution–pooling blocks, a flattening layer, and a configurable dense layer that receives perturbed activations. 
A function \textit{modelStrategyAdder()} enabled the interchangeable integration of Dropout, Gaussian Dropout, PerNodeDrop variants, DropConnect, or MaskEnsemble, allowing flexible control over unit count and drop rate. 
The network was concluded with a 64-unit ReLU layer and a 10-class softmax output, trained using the Adam optimizer with categorical cross-entropy loss and accuracy as the performance metrics.
 
Fig.~\ref{fig:cifar10_bar} summarizes the best configurations, whereas Fig.~\ref{fig:cifar_scatter} shows validation and epoch-end training losses, revealing the convergence and generalization behavior of each method.

\begin{figure}[htbp]
    \centering
    \includegraphics[width=0.48\textwidth]{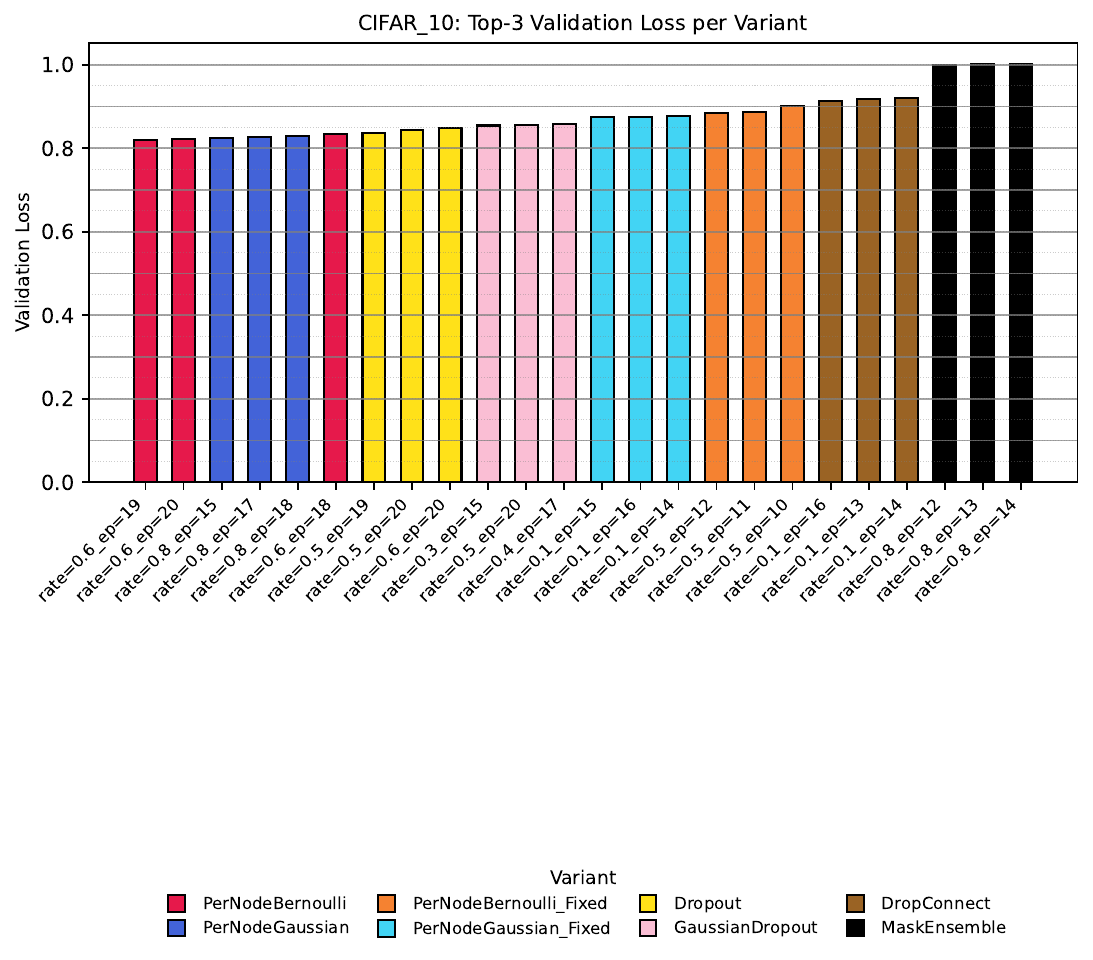}
    \caption{Top-3 best validation-loss configurations per variant on the CIFAR-10 dataset. 
    Each bar represents the configuration $(\text{variant}, \text{drop\_rate}, \text{epoch})$ 
    achieving the competitive validation loss for that regularization method.}
    \label{fig:cifar10_bar}
\end{figure}

\begin{figure}[htbp]
    \centering
    \includegraphics[width=0.48\textwidth]{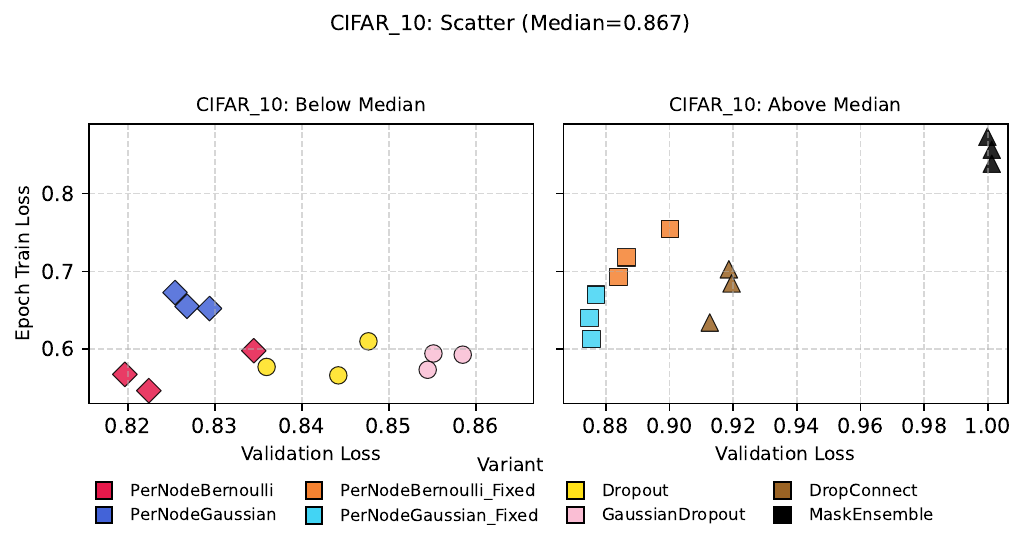}
    \caption{Validation loss versus epoch-end training loss for the top-3 configurations of each regularization variant on the CIFAR-10 dataset. 
    The plot is split at the median validation loss (0.878) to distinguish relatively better- and poorer-generalizing configurations. 
    Each point represents one of the best-performing setups identified in Fig.~\ref{fig:cifar10_bar}, color- and shape-coded by variant. 
    The left panel highlights well-regularized models with low validation and training losses, whereas the right panel corresponds to higher-loss configurations.}
    \label{fig:cifar_scatter}
\end{figure}

\begin{table}[htbp]
\centering
\caption{Top-3 configurations per regularization variant on CIFAR-10.
Each entry reports validation and epoch-end training metrics for the best runs visualized in Fig.~\ref{fig:cifar10_bar} and Fig.~\ref{fig:cifar_scatter}. 
The suffix ``\_F'' denotes the \textit{fixed perturbation} variant, where a static mask was used instead of dynamic stochastic sampling. 
Columns \textbf{DR} and \textbf{Ep} represent the drop rate (or the corresponding hyperparameter) and the training epoch, respectively. 
\textbf{V-Acc} and \textbf{V-Loss} indicate validation accuracy and validation loss, whereas \textbf{T-Loss} and \textbf{T-Acc} denote the epoch-end training loss and accuracy computed on training data that was not subjected to the perturbing function.}
\label{tab:cifar10_results}
\renewcommand{\arraystretch}{1.05}
\setlength{\tabcolsep}{2.5pt}
\scriptsize
\begin{tabular}{lcccccc}
\hline
\textbf{Variant} & \textbf{DR} & \textbf{Ep} &
\textbf{V-Acc} & \textbf{V-Loss} &
\textbf{T-Loss} & \textbf{T-Acc} \\ 
\hline

PerNodeBernoulli & 0.6 & 19 & .72 & .82 & .567 & .808 \\
PerNodeBernoulli & 0.6 & 20 & .721 & .822 & .546 & .811 \\
PerNodeGaussian & 0.8 & 15 & .713 & .825 & .673 & .768 \\
PerNodeGaussian & 0.8 & 17 & .713 & .827 & .655 & .774 \\
PerNodeGaussian & 0.8 & 18 & .715 & .829 & .652 & .774 \\
PerNodeBernoulli & 0.6 & 18 & .716 & .834 & .598 & .794 \\
Dropout & 0.5 & 19 & .708 & .836 & .577 & .805 \\
Dropout & 0.5 & 20 & .709 & .844 & .566 & .807 \\
Dropout & 0.6 & 20 & .703 & .848 & .61 & .792 \\
GaussianDropout & 0.3 & 15 & .711 & .854 & .573 & .804 \\
GaussianDropout & 0.5 & 20 & .705 & .855 & .594 & .8 \\
GaussianDropout & 0.4 & 17 & .705 & .858 & .593 & .797 \\
PerNodeGaussian\_F & 0.1 & 15 & .702 & .875 & .64 & .783 \\
PerNodeGaussian\_F & 0.1 & 16 & .704 & .876 & .613 & .792 \\
PerNodeGaussian\_F & 0.1 & 14 & .701 & .877 & .67 & .773 \\
PerNodeBernoulli\_F & 0.5 & 12 & .692 & .884 & .693 & .765 \\
PerNodeBernoulli\_F & 0.5 & 11 & .691 & .886 & .718 & .757 \\
PerNodeBernoulli\_F & 0.5 & 10 & .689 & .9 & .755 & .744 \\
DropConnect & 0.1 & 16 & .701 & .913 & .634 & .782 \\
DropConnect & 0.1 & 13 & .693 & .919 & .703 & .757 \\
DropConnect & 0.1 & 14 & .694 & .919 & .684 & .763 \\
MaskEnsemble & 0.8 & 12 & .652 & 1.000 & .873 & .698 \\
MaskEnsemble & 0.8 & 13 & .655 & 1.001 & .857 & .705 \\
MaskEnsemble & 0.8 & 14 & .656 & 1.001 & .838 & .711 \\

\hline
\end{tabular}
\end{table}

\subsection{Text Domain: RCV1-v2 Dataset}

The \textbf{RCV1-v2 dataset}~\cite{RCV_V2_Lewis_2004}—a multi-label text classification benchmark was used to evaluate the proposed regularization schemes in a non-visual domain. 
Precomputed TF-IDF feature vectors were utilized, retaining the 25 most frequent categories, resulting in 23,149 documents and 47,236 features. 
Data was divided into 80\% training and 20\% validation subsets.

A feedforward neural network was implemented with an input dimensionality matching the TF–IDF feature space. The architecture comprised fully connected layers with 1024, 256, and 128 units, with a configurable regularization module (\textit{modelStrategyAdder}) inserted between layers to apply Dropout, Gaussian Dropout, PerNodeDrop variants, or DropConnect. The output layer consisted of 25 sigmoid-activated neurons corresponding to the top 25 topics in the RCV1-v2 corpus, with each neuron estimating the probability of its associated label. The model was trained using the Adam optimizer with binary cross-entropy loss, which is appropriate for the multilabel nature of the dataset.

The \textbf{MaskEnsemble} variant was omitted, as the TF-IDF feature dimensionality was not compatible with the required mask-block partitioning.

Fig.~\ref{fig:nlp_bar} and Fig.~\ref{fig:nlp_scatter} show the comparative performance of the top-three configurations per variant, analogous to the CIFAR-10 plots in Fig.~\ref{fig:cifar10_bar} and Fig.~\ref{fig:cifar_scatter}. 
The fixed perturbation variants of PerNodeDrop achieved the lowest validation losses and fastest convergence, typically within five epochs, whereas the dynamic variants maintained slightly higher validation losses, indicating a stronger regularization effect that may improve robustness.This suggests that deterministic masking is more effective for high-dimensional, sparse TF--IDF representations, whereas stochastic masking provides marginal regularization gains.

\begin{figure}[htbp]
    \centering
    \includegraphics[width=0.48\textwidth]{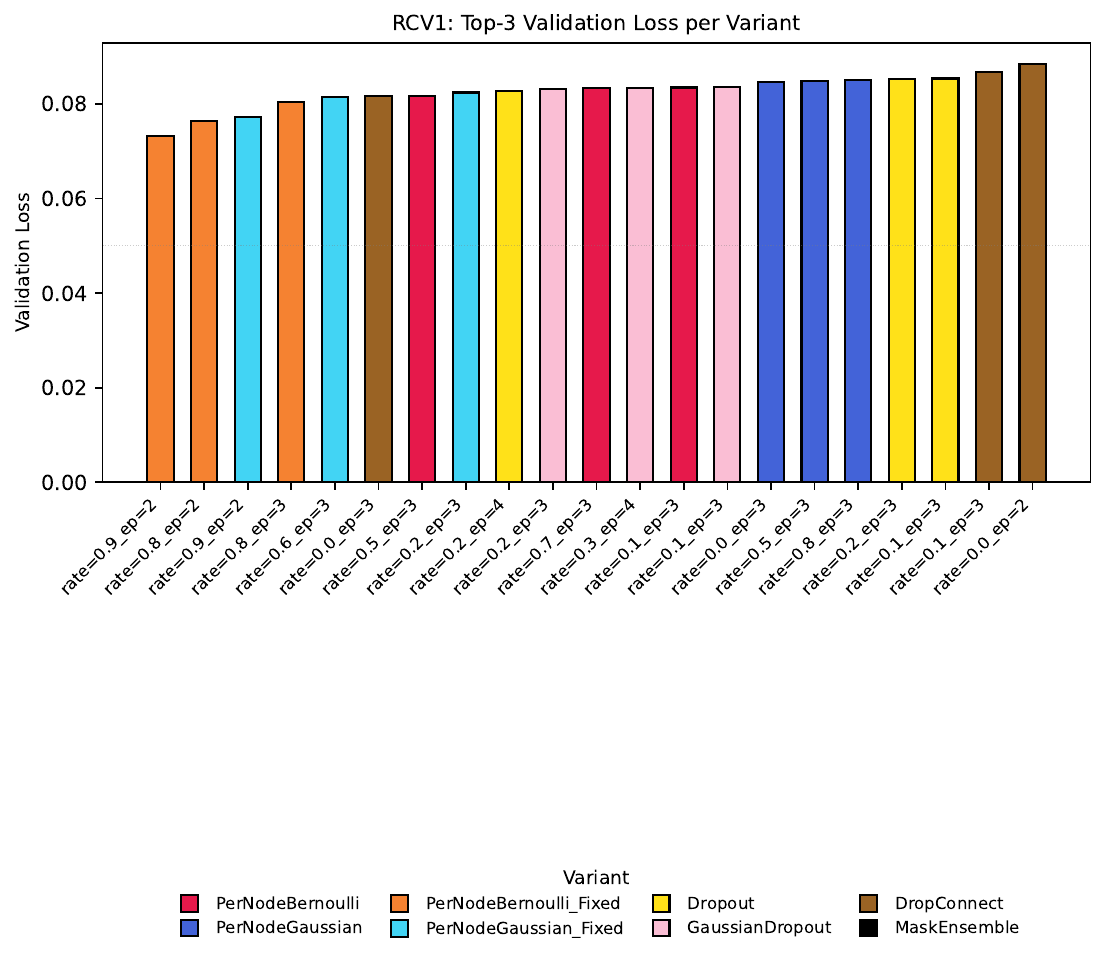}
    \caption{Top-3 best validation-loss configurations per variant on the 20 RCV1-v2 dataset, equivalent to Fig.~\ref{fig:cifar10_bar} for CIFAR-10. 
    Each bar represents the configuration $(\text{variant}, \text{drop\_rate}, \text{epoch})$ 
    achieving the lowest validation loss for that regularization method.}
    \label{fig:nlp_bar}
\end{figure}

\begin{figure}[htbp]
    \centering
    \includegraphics[width=0.48\textwidth]{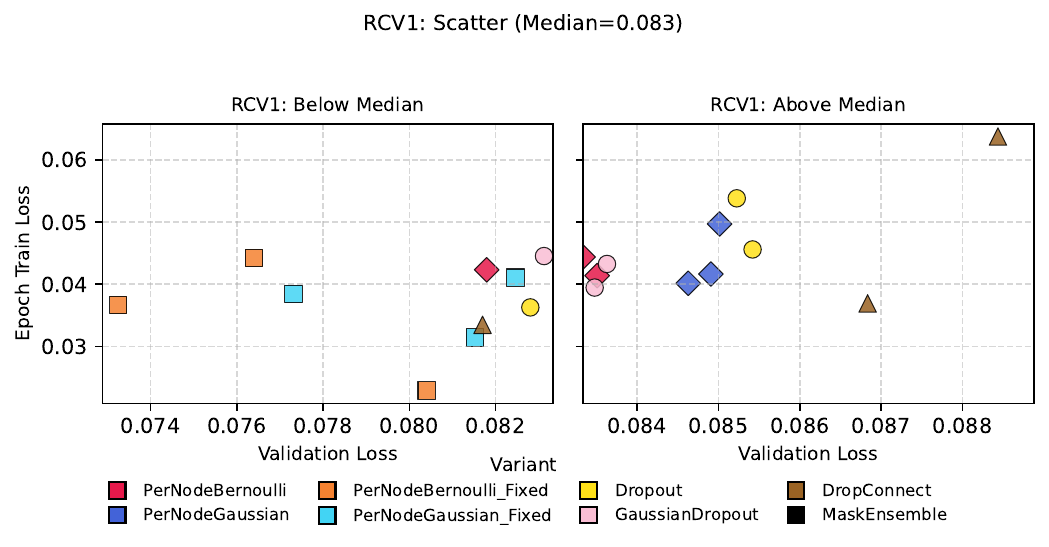}
    \caption{Validation loss versus epoch-end training loss for the top-3 configurations of each regularization variant on the 20 RCV1-v2 dataset, corresponding to Fig.~\ref{fig:cifar_scatter} for CIFAR-10. 
    Each point denotes one of the best-performing setups, color- and shape-coded by variant. 
    The split at the median validation loss (0.083) differentiates relatively better- and poorer-generalizing configurations.}
    \label{fig:nlp_scatter}
\end{figure}

\begin{table}[htbp]
\centering
\caption{Top-3 configurations per regularization variant on the 20 RCV1-v2 dataset.
Each entry reports validation and epoch-end training metrics for the best runs visualized in Fig.~\ref{fig:nlp_bar} and Fig.~\ref{fig:nlp_scatter}. 
The suffix ``\_F'' denotes the \textit{fixed perturbation} variant, where a static mask was used instead of dynamic stochastic sampling. 
Columns \textbf{DR} and \textbf{Ep} represent the drop rate (or corresponding hyperparameter) and the training epoch, respectively. 
\textbf{V-Acc} and \textbf{V-Loss} indicate validation accuracy and validation loss, whereas \textbf{T-Loss} and \textbf{T-Acc} denote the epoch-end training loss and accuracy computed on training data that was not subjected to the perturbing function.}
\label{tab:nlp_results}
\renewcommand{\arraystretch}{1.05}
\setlength{\tabcolsep}{2.5pt}
\scriptsize
\begin{tabular}{lcccccc}
\hline
\textbf{Variant} & \textbf{DR} & \textbf{Ep} &
\textbf{V-Acc} & \textbf{V-Loss} &
\textbf{T-Loss} & \textbf{T-Acc} \\ 
\hline
PerNodeBernoulli\_F     & 0.9 & 2 & .810 & .073 & .037 & .831 \\
PerNodeBernoulli\_F     & 0.8 & 2 & .845 & .076 & .044 & .870 \\
PerNodeGaussian\_F      & 0.9 & 2 & .811 & .077 & .038 & .830 \\
PerNodeBernoulli\_F     & 0.8 & 3 & .816 & .080 & .023 & .839 \\
PerNodeGaussian\_F      & 0.6 & 3 & .838 & .082 & .031 & .876 \\
DropConnect             & 0.0 & 3 & .857 & .082 & .033 & .893 \\
PerNodeBernoulli        & 0.5 & 3 & .802 & .082 & .042 & .830 \\
PerNodeGaussian\_F      & 0.2 & 3 & .825 & .082 & .041 & .850 \\
Dropout                 & 0.2 & 4 & .838 & .083 & .036 & .871 \\
GaussianDropout         & 0.2 & 3 & .864 & .083 & .045 & .894 \\
PerNodeBernoulli        & 0.7 & 3 & .828 & .083 & .044 & .849 \\
GaussianDropout         & 0.3 & 4 & .829 & .083 & .039 & .857 \\
PerNodeBernoulli        & 0.1 & 3 & .845 & .084 & .041 & .879 \\
GaussianDropout         & 0.1 & 3 & .833 & .084 & .043 & .858 \\
PerNodeGaussian         & 0.0 & 3 & .832 & .085 & .040 & .860 \\
PerNodeGaussian         & 0.5 & 3 & .847 & .085 & .042 & .885 \\
PerNodeGaussian         & 0.8 & 3 & .813 & .085 & .050 & .840 \\
Dropout                 & 0.2 & 3 & .839 & .085 & .054 & .867 \\
Dropout                 & 0.1 & 3 & .846 & .085 & .046 & .878 \\
DropConnect             & 0.1 & 3 & .806 & .087 & .037 & .831 \\
DropConnect             & 0.0 & 2 & .862 & .088 & .064 & .884 \\
\hline
\end{tabular}
\end{table}

\subsection{Audio Domain: Google Voice Spectrogram}

The Mini Speech Commands dataset \textbf{TensorFlow Speech Commands, 2020}~\cite{Warden2018SpeechCommands} was used to evaluate the models on the audio data. Each .wav file was converted into a 128 × 128 Mel-spectrogram using librosa, normalized to zero-mean and unit variance, and labeled according to its spoken command category. The dataset was divided into 7:3  training and validation subsets. A CNN model with two convolution–pooling blocks and a flattening layer was used, followed by a configurable regularization module (modelStrategyAdder) that allows interchangeable use of Dropout, Gaussian Dropout, PerNodeDrop variants, DropConnect, or MaskEnsemble. The network ends with a 64-unit ReLU layer and a softmax output across all spoken classes, trained with the Adam optimizer using categorical cross-entropy loss and accuracy as the performance metrics.

\begin{figure}[htbp]
    \centering
    \includegraphics[width=0.48\textwidth]{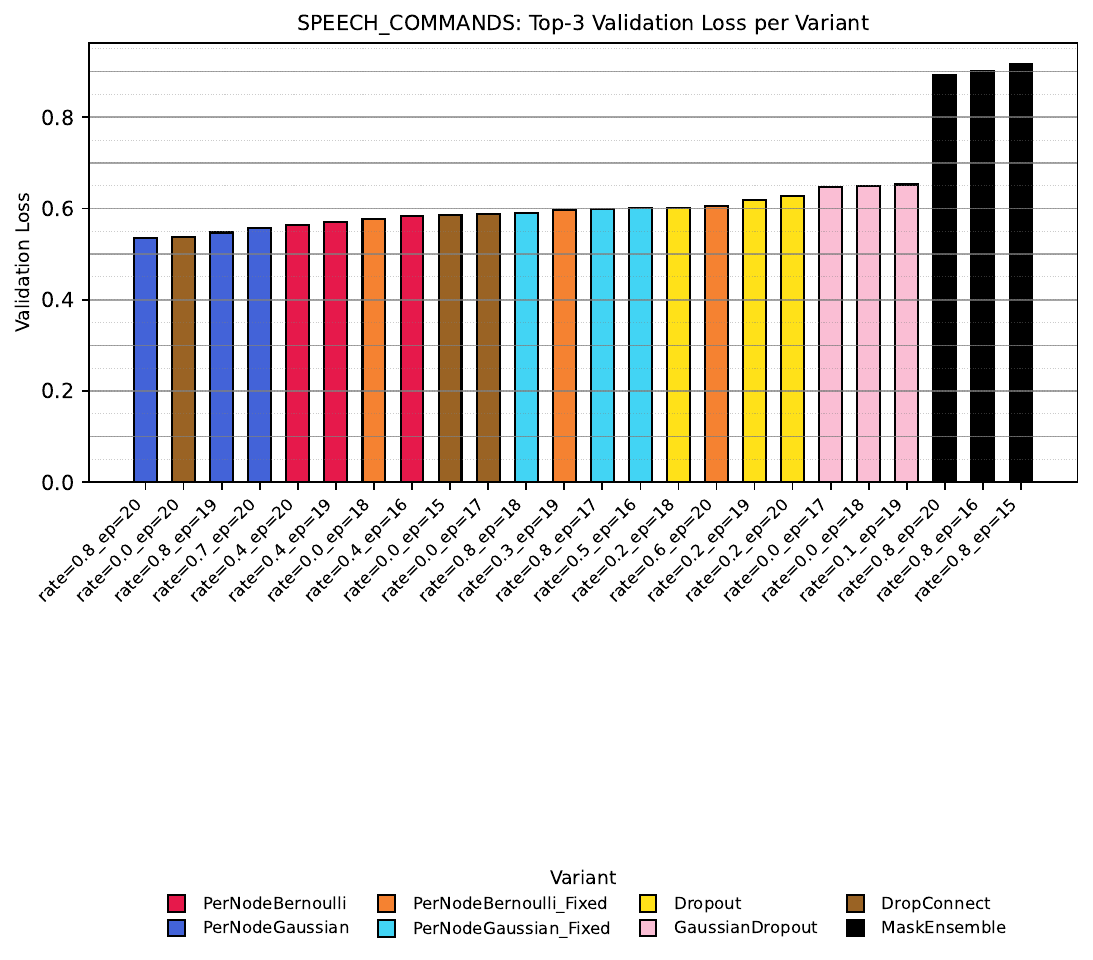}
    \caption{Top-3 best validation-loss configurations per variant on the Mini Speech Commands (Spectrogram) dataset. 
    Each bar represents the configuration $(\text{variant}, \text{drop\_rate}, \text{epoch})$ 
    achieving the lowest validation loss for that regularization method.}
    \label{fig:spectro_bar}
\end{figure}

\begin{figure}[htbp]
    \centering
    \includegraphics[width=0.48\textwidth]{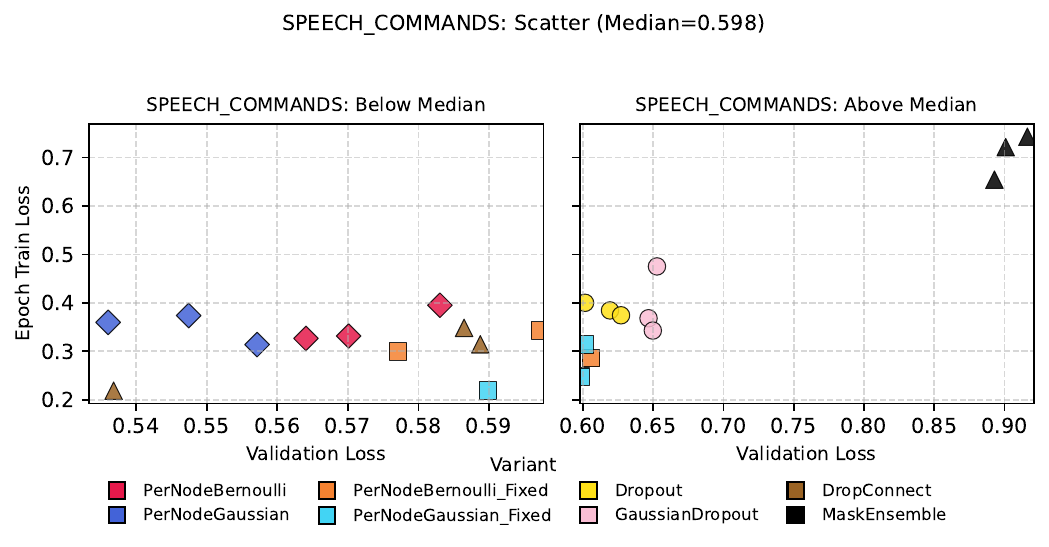}
    \caption{Validation loss versus epoch-end training loss for the top-3 configurations of each regularization variant on the Mini Speech Commands dataset. 
    The plot is split at the median validation loss (0.607) to distinguish relatively better- and poorer-generalizing configurations. 
    Each point represents one of the best-performing setups identified in Fig.~\ref{fig:spectro_bar}, color- and shape-coded by variant. 
    The left panel highlights well-regularized models with low validation and training losses, whereas the right panel corresponds to higher-loss configurations.}
    \label{fig:spectro_scatter}
\end{figure}

\begin{table}[htbp]
\centering
\caption{Top-3 configurations per regularization variant on the Mini Speech Commands (Spectrogram) dataset.
Each entry reports validation and epoch-end training metrics for the best runs visualized in Fig.~\ref{fig:spectro_bar} and Fig.~\ref{fig:spectro_scatter}.
The suffix ``\_F'' denotes the \textit{fixed perturbation} variant, where a static mask was used instead of dynamic stochastic sampling. 
Columns \textbf{DR} and \textbf{Ep} represent the drop rate (or the corresponding hyperparameter) and the training epoch, respectively. 
\textbf{V-Acc} and \textbf{V-Loss} indicate validation accuracy and validation loss, whereas \textbf{T-Loss} and \textbf{T-Acc} denote the epoch-end training loss and accuracy computed on training data that was not subjected to the perturbing function.}

\label{tab:spectro_results}
\renewcommand{\arraystretch}{1.05}
\setlength{\tabcolsep}{2.5pt}
\scriptsize
\begin{tabular}{lcccccc}
\hline
\textbf{Variant} & \textbf{DR} & \textbf{Ep} &
\textbf{V-Acc} & \textbf{V-Loss} &
\textbf{T-Loss} & \textbf{T-Acc} \\ 
\hline
PerNodeGaussian         & 0.8 & 20 & .815 & .536 & .360 & .889 \\
DropConnect             & 0.0 & 20 & .833 & .537 & .219 & .935 \\
PerNodeGaussian         & 0.8 & 19 & .807 & .547 & .374 & .881 \\
PerNodeGaussian         & 0.7 & 20 & .815 & .557 & .314 & .907 \\
PerNodeBernoulli        & 0.4 & 20 & .809 & .564 & .327 & .899 \\
PerNodeBernoulli        & 0.4 & 19 & .809 & .570 & .331 & .894 \\
PerNodeBernoulli\_F     & 0.0 & 18 & .802 & .577 & .300 & .907 \\
PerNodeBernoulli        & 0.4 & 16 & .804 & .583 & .395 & .877 \\
DropConnect             & 0.0 & 15 & .798 & .586 & .348 & .889 \\
DropConnect             & 0.0 & 17 & .807 & .589 & .314 & .900 \\
PerNodeGaussian\_F      & 0.8 & 18 & .812 & .590 & .219 & .936 \\
PerNodeBernoulli\_F     & 0.3 & 19 & .797 & .597 & .342 & .891 \\
PerNodeGaussian\_F      & 0.8 & 17 & .803 & .598 & .247 & .925 \\
PerNodeGaussian\_F      & 0.5 & 16 & .798 & .601 & .314 & .902 \\
Dropout                 & 0.2 & 18 & .788 & .602 & .400 & .869 \\
PerNodeBernoulli\_F     & 0.6 & 20 & .800 & .606 & .286 & .911 \\
Dropout                 & 0.2 & 19 & .779 & .619 & .384 & .871 \\
Dropout                 & 0.2 & 20 & .780 & .627 & .374 & .875 \\
GaussianDropout         & 0.0 & 17 & .777 & .647 & .368 & .886 \\
GaussianDropout         & 0.0 & 18 & .780 & .650 & .343 & .893 \\
GaussianDropout         & 0.1 & 19 & .768 & .653 & .475 & .839 \\
MaskEnsemble            & 0.8 & 20 & .701 & .893 & .654 & .774 \\
MaskEnsemble            & 0.8 & 16 & .686 & .901 & .721 & .747 \\
MaskEnsemble            & 0.8 & 15 & .678 & .916 & .743 & .732 \\
\hline
\end{tabular}
\end{table}

\subsection{Execution time}
\begin{table}[h!]
\centering
\caption{Mean Epoch Duration (seconds) Across Datasets and Regularization Variants}
\label{tab:epoch_time}
\begin{tabular}{l l c c}
\hline
\textbf{Dataset} & \textbf{Variant} & \textbf{Mean (s)} & \textbf{Std. Dev.} \\ 
\hline
\multirow{8}{*}{CIFAR--10}
 & DropConnect               & 3.07 & 0.27 \\
 & MaskEnsemble              & 3.01 & 0.20 \\
 & PerNodeBernoulli          & 4.43 & 0.28 \\
 & PerNodeBernoulli\_Fixed   & 3.13 & 0.32 \\
 & PerNodeGaussian           & 5.27 & 0.27 \\
 & PerNodeGaussian\_Fixed    & 3.20 & 0.47 \\
 & Dropout                   & 3.04 & 0.27 \\
 & GaussianDropout           & 3.04 & 0.27 \\
\hline
\multirow{8}{*}{Speech Commands}
 & DropConnect               & 1.59 & 0.12 \\
 & MaskEnsemble              & 1.60 & 0.17 \\
 & PerNodeBernoulli          & 3.06 & 0.11 \\
 & PerNodeBernoulli\_Fixed   & 1.70 & 0.18 \\
 & PerNodeGaussian           & 3.50 & 0.14 \\
 & PerNodeGaussian\_Fixed    & 1.70 & 0.13 \\
 & Dropout                   & 1.60 & 0.15 \\
 & GaussianDropout           & 1.59 & 0.11 \\
\hline
\multirow{7}{*}{RCV1--v2}
 & DropConnect               & 9.49 & 0.79 \\
 & PerNodeBernoulli          & 9.53 & 0.49 \\
 & PerNodeBernoulli\_Fixed   & 9.47 & 0.47 \\
 & PerNodeGaussian           & 9.52 & 0.49 \\
 & PerNodeGaussian\_Fixed    & 9.50 & 0.46 \\
 & Dropout                   & 9.49 & 0.49 \\
 & GaussianDropout           & 9.46 & 0.52 \\
\hline
\end{tabular}
\end{table}

\paragraph{Epoch duration and computational profile.}
Table~\ref{tab:epoch_time} summarizes the mean epoch durations recorded for each regularization variant across the three datasets. 
While RCV1--v2 exhibited almost identical timings for all methods  ($\approx$9.5\,s per epoch), the CIFAR--10 and Speech Commands datasets revealed 
moderate overheads for the PerNodeDrop variants.  Specifically, PerNodeBernoulli and PerNodeGaussian required approximately 1.3$\times$ and 2$\times$ the baseline time for CIFAR--10 and Speech Commands, respectively.  This difference arises because the spectrogram input (128$\times$128) is of eight times more spatial resolution than that of the CIFAR--10 image (32$\times$32$\times$3). Even after two pooling stages, the flattened feature representation in the spectrogram model remains $\sim$65k elements, compared to $\sim$4k in CIFAR--10. This delay is primarily due to the per-sample mask generation and element-wise multiplications in the dense layers. Since the convolutional blocks dominate the total computation and benefit from GPU parallelism, the overall runtime increases sub-linearly with input size, remaining within a practically acceptable range. Hence, the observed slowdown does not imply inefficiency in the PerNodeDrop formulation but reflects the proportional increase in dense-layer connectivity for high-dimensional feature maps.

\subsection{Summary of Experiments Outcome}

Evaluation of model overfitting should consider not only the minimum validation loss but also the similarity between training and validation losses under identical inference conditions.

In all experiments, the \textbf{PerNodeDrop} variants demonstrated the most favorable validation loss while keeping the training loss relatively close to the training loss.
As shown in Fig.~\ref{fig:cifar_scatter} and Fig.~\ref{fig:spectro_scatter}, corresponding to the CIFAR-10 and spectrogram-based audio datasets, both the Bernoulli and Gaussian forms of PerNodeDrop achieved the lowest validation losses while maintaining training losses that were comparable to—or slightly higher than—those of the baseline regularizers. The close proximity between validation and training losses, even when the validation loss was minimal, indicates stronger resistance to overfitting and improved generalization.
In contrast, for the text-domain RCV1-v2 dataset (Fig.~\ref{fig:nlp_bar} and Fig.~\ref{fig:nlp_scatter}), the \textbf{Fixed Perturbation} variants of PerNodeDrop outperformed their dynamic counterparts, converging in fewer than five epochs.
This behavior suggests that fixed masking may be better suited for high-dimensional, sparse TF–IDF features, or for datasets that may tend to converge faster, where stochastic noise offers limited representational
diversity.

\section{Discussion}

\begin{table}[htbp]
\centering
\caption{Mean rank per regularization variant computed from the top-5 configurations with the lowest validation loss across all epochs and drop rates, grouped by variant. Lower values indicate better performance.}
\label{tab:mean_rank_per_dataset}
\renewcommand{\arraystretch}{1.1}
\setlength{\tabcolsep}{6pt}
\scriptsize
\begin{tabular}{lc}
\multicolumn{2}{l}{\textbf{Dataset: CIFAR-10}} \\
\hline
\textbf{Variant} & \textbf{Mean Rank} \\
\hline
PerNodeGaussian            &  5.4 \\
PerNodeBernoulli           &  7.0 \\
Dropout                    & 11.6 \\
GaussianDropout            & 18.0 \\
PerNodeGaussian\_Fixed     & 23.0 \\
PerNodeBernoulli\_Fixed    & 28.0 \\
DropConnect                & 33.0 \\
MaskEnsemble               & 38.0 \\
\hline
\multicolumn{2}{l}{\textit{Friedman} $\chi^2 = 34.600$, $p = 0.00001$} \\
\multicolumn{2}{l}{Kendall’s $W = 0.989$ (high concordance)} \\
\\[-4pt]
\multicolumn{2}{l}{\textbf{Dataset: RCV1-v2}} \\
\hline
\textbf{Variant} & \textbf{Mean Rank} \\
\hline
PerNodeBernoulli\_Fixed    &  3.6 \\
PerNodeGaussian\_Fixed     &  8.6 \\
PerNodeBernoulli           & 16.8 \\
GaussianDropout            & 18.8 \\
PerNodeGaussian            & 24.4 \\
Dropout                    & 25.4 \\
DropConnect                & 28.4 \\
\hline
\multicolumn{2}{l}{\textit{Friedman} $\chi^2 = 25.114$, $p = 0.00033$} \\
\multicolumn{2}{l}{Kendall’s $W = 0.837$ (high concordance)} \\
\\[-4pt]
\multicolumn{2}{l}{\textbf{Dataset: Speech Commands}} \\
\hline
\textbf{Variant} & \textbf{Mean Rank} \\
\hline
PerNodeGaussian            &  4.6 \\
PerNodeBernoulli           &  9.4 \\
DropConnect                & 12.4 \\
PerNodeBernoulli\_Fixed    & 19.0 \\
PerNodeGaussian\_Fixed     & 20.6 \\
Dropout                    & 27.8 \\
GaussianDropout            & 32.2 \\
MaskEnsemble               & 38.0 \\
\hline
\multicolumn{2}{l}{\textit{Friedman} $\chi^2 = 34.467$, $p = 0.00001$} \\
\multicolumn{2}{l}{Kendall’s $W = 0.985$ (high concordance)} \\
\hline
\end{tabular}
\end{table}

The experiments conducted across three distinct datasets —vision (CIFAR-10), text (RCV1-v2), and audio -spectrograms (Mini Speech Commands)—demonstrate that the proposed \textit{PerNodeDrop} variants competitively outperform or match the performance of classical regularization methods such as Dropout, Gaussian Dropout, DropConnect, and MaskEnsemble. The validation and epoch-end training loss scatter plots (Figs.~\ref{fig:cifar_scatter},~\ref{fig:nlp_scatter},~\ref{fig:spectro_scatter}) and their corresponding summary tables (Tables~\ref{tab:cifar10_results},~\ref{tab:nlp_results},~\ref{tab:spectro_results}) show that \textit{PerNodeDrop} maintains a closer proximity between training and validation losses, indicating a more stable and generalizable learning behavior. Observed behavior of DropConnect at extreme zero drop rates on the RCV1-v2 dataset (see Table~\ref{tab:nlp_results} and Fig.~\ref{fig:nlp_scatter}) may reflect implementation-level scaling effects rather than inherent methodological superiority.

Across datasets, the \textit{PerNodeGaussian} and \textit{PerNodeBernoulli} dynamic variants achieved the lowest mean validation losses (Tables~\ref{tab:cifar10_results}, and ~\ref{tab:spectro_results}).  

In the vision and audio domains, the optimal drop rate was approximately $\approx 0.4$--$0.6$; higher rates yielded negligible gains or slight performance degradation, indicating a regularization strength that mitigates co-adaptation while preserving informative activations.
In RCV1-v2, fixed-mask variants outperformed dynamic masks, likely because temporally consistent perturbations encourage stable feature reuse in sparse, high-dimensional TF--IDF representations. This observation agrees with that of Srivastava et~al.~\cite{srivastava2014dropout}, who reported that excessive dropout noise can impede the convergence of linguistic features.
Thus, the relative benefit of fixed versus dynamic masking depends on the input-structure characteristics.

Variant rankings were highly concordant across datasets (Kendall's W $>$ 0.83; Friedman $\chi^2$ = 25--35, $p < 0.001$), supporting the robustness of these trends despite moderate effect sizes. The results indicate that finer-grained stochasticity and sample-wise diversity can improve generalization without altering the underlying architecture. PerNodeDrop’s per-sample stochastic masks are intended to regularize while preserving beneficial co-adaptations, effectively acting as a lightweight implicit ensemble. The fixed variant enforces a consistent subnetwork per sample (analogous to MaskEnsemble to some extent), whereas the dynamic variant adds sample-level randomness akin to Bayesian posterior sampling. In-depth and formal uncertainty quantification (e.g., calibration, expected calibration error, negative log-likelihood) was not performed and is left to future work.

Several limitations of this study should be noted. First, the experiments used relatively shallow architectures; the scalability of \textit{PerNodeDrop} to deeper CNNs and Transformer models remains untested. Second, computational efficiency was not evaluated in terms of FLOPs or wall-clock time. Third, the anomalous strong performance of \textit{DropConnect} at zero drop rate may reflect framework-level scaling artifacts and hence requires controlled revalidation. Finally, although the expected-loss analysis (Eq.~\ref{eq:pernode_general}) motivates the regularization mechanism, the explicit contribution of mask variance ($\sigma_{M}^{2}$) to generalization has not been quantified. Third, this study did not perform an exhaustive hyperparameter search for Masksembles.

In summary, \textit{PerNodeDrop} delivered modest yet reasonably consistent improvements in generalization and training stability across modalities. The method is suitable for adoption in research, but broader scalability benchmarks and calibration analyses are needed to establish its reliability under large-scale and perturbed conditions. Hence, while \textit{PerNodeDrop} appears promising as a unified, fine-grained regularization mechanism, its complete reliability evaluation under large-scale and perturbed conditions remains an open and worthwhile direction for future work.

\section{Conclusion}
The proposed \textbf{PerNodeDrop} framework introduces a perturbation mechanism that unifies the benefits of node-level specialization and stochastic regularization. The technique of dropping incoming signals per node is functionally analogous to selectively dropping outgoing weights, as in DropConnect, although the stochastic structures differ in practice. The structural and computational simplicity of \textbf{PerNodeDrop} makes it practical to adapt at the granular level of an individual input instance, unlike DropConnect, which in practice operates at the batch level.

Across diverse domains, including vision (CIFAR-10), text (RCV1-v2), and audio (Mini Speech Commands), the results suggest that PerNodeDrop achieves performance comparable to conventional Dropout, Gaussian Dropout, and DropConnect on validation.

The PerNodeDrop formulation unifies the dynamic masking behavior of Dropout and DropConnect with the subnet ensembling behavior of Masksembles into a single, computationally efficient framework. Incorporating Bernoulli, Gaussian, fixed, and dynamic modes (and extensible to partial-Gaussian forms), PerNodeDrop reduces overfitting while preserving the model’s capacity to learn meaningful features. Across vision, text, and audio tasks, the method consistently exhibits a slightly higher training loss paired with a lower validation loss, indicating improved generalization compared to other noise-based regularizer variants evaluated in this study. The results suggest that node-level stochastic masking provides an effective balance between regularization and model capacity.

The scatter plots (Fig.~\ref{fig:cifar_scatter}, Fig.~\ref{fig:nlp_scatter}, and Fig.~\ref{fig:spectro_scatter}) show that PerNodeDrop variants consistently end with higher training loss and frequently achieve the lowest validation loss among the compared methods. This combination of elevated training loss alongside comparatively strong validation performance reflects effective regularization and reduced overfitting. A plausible explanation is that the node-level perturbations reduce excessive co-adaptation while still permitting valuable feature interactions, encouraging neurons to learn complementary representations without limiting overall learning capacity.

The fixed perturbation variants (\textit{PerNodeDrop-Fixed}) behave analogously to Maskensemble of shared subnetworks that repeatedly sample from the same structural mask. As seen in the case of high-dimensional but fast-converging TF–IDF experiments, (\textit{PerNodeDrop-Fixed}) was efficient; in most scenarios, the dynamic nature of \textit{PerNodeDrop} was found to be advantageous. 

\section{Future Work}
The dynamic masking variants induce sample-specific perturbations that can be viewed as implicit ensembling, thereby suggesting potential advantages for uncertainty estimation, in line with the interpretations of Kendall and Gal~\cite{Kendall2017}. These interpretations are presented as structural analogies, and this study does not claim validated Bayesian uncertainty estimation. Consequently, calibration metrics (ECE, NLL), OOD detection performance, and Bayesian perspectives of the induced stochasticity represent natural directions for future work. Future work will also investigate second-order optimization characteristics by analyzing gradient norms and curvature-based diagnostics, such as Hessian-informed measures. In addition, extending PerNodeDrop to convolutional and transformer-based architectures (e.g., ResNet, MobileNetV2, and Transformers) and studying its interaction with attention mechanisms and self-distillation frameworks may provide deeper insights into its behavior in large-scale models.

 Conventional dropout disrupts temporal consistency across time steps in sequential architectures such as RNNs and LSTMs.  To maintain temporal coherence, \textit{variational dropout}~\cite{Gal2016} was devised to enforce a fixed dropout mask over all timesteps of a sequence.  This approach also preserves and propagates the same mask, which adds computational complexity, memory cost, and stochastic variation during training.
 The conceptual similarity between dynamic per-node masking in \textit{PerNodeDrop} and timestep-consistent variational dropout perturbations allows for a hybrid formulation. Dynamically evolving yet temporally correlated PerNode masks combine the stability of variational dropout with the adaptive diversity of \textit{PerNodeDrop} in this design. This may enable sequence models to capture uncertainty and regularize without the computational cost of explicit variational distributions.

\section*{Reproducibility Statement}
All datasets used in this study, CIFAR-10, RCV1, and Mini Speech Commands, are publicly available. 
The experiments were conducted using Google Colab TensorFlow~2.12 on an NVIDIA RTX-series GPU with batch size 128-256 and learning rate~1e$^{-3}$. 
The source code implementing PerNodeDrop, along with configuration files and training logs, will be publicly released upon acceptance and publication of this work. 

\section*{Author Contribution}
Gelesh G. Omathil was solely responsible for all aspects of this research. This included conceptualization, development of the  research methodology, research design, data curation, experimental  investigation, implementation, validation, formal analysis, and interpretation of results. The author also undertook the preparation, writing, and revision of the manuscript, including the development of figures, tables, and supporting materials. The author reviewed and approved the final version of the manuscript and takes full responsibility for the integrity and accuracy of the work.

No other individual contributed to the research or manuscript preparation, and no independent human peer review was conducted prior to submission.

\section*{Acknowledgment}
This research was conducted as part of Gelesh G. Omathil’s academic research and received partial financial support from Chemophilic Data Sage LLP. Dr. Sreeja C. S. was formally assigned as the research supervisor.

\newpage

\appendices

\end{document}